\documentclass[sigplan,10pt]{acmart}
\startPage{1}

\setcopyright{none}

\usepackage{booktabs}   
\usepackage{subcaption} 
\usepackage{array}   
\newcolumntype{L}{>{$}l<{$}} 
\newcolumntype{T}{>{\centering\arraybackslash}m{3cm}}

\begin{document}

\title{Learning to Combine Instructions in LLVM Compiler}         


\author{Sandya Mannarswamy}
\affiliation{
  \institution{Intel India}            
  \city{Bangalore}
  \country{India}                    
}
\email{sandya.mannarswamy@intel.com}          

\author{Dibyendu Das}
\affiliation{
  \institution{Intel India}           
  \city{Bangalore}
  \country{India}       
}
\email{dibyendu.das@intel.com}

\begin{abstract}
Instruction combiner (IC) is a critical compiler optimization pass, which replaces a sequence of instructions with an equivalent and optimized instruction sequence at basic block level. There can be thousands of instruction-combining patterns which need to be frequently updated as new coding idioms/applications and novel hardware evolve over time. This results in frequent updates to the IC optimization pass thereby incurring considerable human effort and high software maintenance costs. To mitigate these challenges associated with the traditional IC, we design and implement a Neural Instruction Combiner (NIC) and demonstrate its feasibility by integrating it into the standard LLVM compiler optimization pipeline. 

NIC leverages neural sequence-to-sequence (Seq2Seq) models for generating optimized encoded IR sequence from the unoptimized encoded IR sequence. To the best of our knowledge, ours is the first work demonstrating the feasibility of a neural instruction combiner built into a full-fledged compiler pipeline. Given the novelty of this task, we built a new dataset for training our NIC neural model. We show that NIC achieves exact match results percentage of 72\% for optimized sequences as compared to traditional IC and neural machine translation metric Bleu precision score of 0.94, demonstrating its feasibility in a production compiler pipeline.
\end{abstract}


\maketitle

\section{Introduction}
Of late, considerable strides have been made in applying deep learning (DL) techniques to software engineering itself, including source code assistance, automatic source code generation and in building software tools ~\cite{Le2020}. The emergence of open-source community software development and large code repositories such as GitHub have accelerated interest in applying DL techniques to programming, compiler optimizations, code generation etc. Neural models have been developed for source code ~\cite{Alon2019} and intermediate code representations ~\cite{VenkataKeerthy2020}. ML models have been used for cost prediction and heuristics selection in compiler optimizations ~\cite{Leather2020}. 

Instruction Combining pass is a basic compiler optimization pass present in all compilers. Instruction Combiner [IC] does local instruction level optimizations on basic blocks (BB) which are jump-free sequential lists of instructions. IC operates on the compiler’s Intermediate Representation (IR) and replaces a sequence of one or more instructions with an optimized and semantically equivalent instruction sequence.  

ICs are typically developed with considerable human effort. There are thousands of patterns that are considered for replacement and these patterns continually need to be added/updated/removed as new coding idioms/applications and new hardware with more sophisticated instruction set architecture (ISA) evolve over time. A typical IC pass often spans several thousand lines of code making it complex to maintain/debug/enhance. Empirical studies show that IC is the most frequently updated pass in the LLVM compiler~\cite{Zhou2021}.

Given the code complexity, software maintenance effort and its wide usage, IC is an ideal target for improvement with machine learnt models. However there also exist considerable challenges in replacing a deterministic and human-written IC with a probabilistic machine learnt NIC. These challenges include representation of the input instruction sequence to the neural model, ensuring correctness of the probabilistic generated code by the neural model, integration of the neural model into a standard compiler optimization pipeline etc. This brings up the question of whether it is feasible to replace traditional IC by a neural model.	

In this paper, we design and implement a Neural Instruction Combiner (NIC) and demonstrate its feasibility by integrating it into the standard LLVM compiler optimization pipeline which can generate executable machine code.  NIC leverages neural Seq2Seq model techniques~\cite{Bahdanau2014} for generating optimized encoded IR sequence from the unoptimized encoded IR sequence at the basic block level, modelling it as monolingual machine translation task. We improve the standard attention mechanism~\cite{Vaswani2017} in Seq2Seq models with a compiler guided attention approach.

NIC consists of three major components 
\begin{itemize}
\item \textbf{NIC Inputter} - This is a compiler module which creates a distilled representation of the IR instruction sequence corresponding to each BB. 
\item \textbf {NIC Converter} - This is a machine learnt model which takes the output of NIC Inputter and converts it to an equivalent optimized sequence.
\item \textbf{NIC Outputter} - This is a compiler module which takes as input the optimized sequence generated by NIC Converter and the original list of IR instructions corresponding to that BB and recreates the standard LLVM IR optimized instruction sequence for that BB. It also performs a set of IR verification checks and translation validity checking using Alive2 tool~\cite{alive22021}.
\end {itemize}

Given the novelty of this task, we create a new dataset for training our NIC neural model. We show that NIC can achieve exact match of 72\% on optimized sequences as compared to traditional IC and Bleu precision score (a standard metric for neural machine translation quality which indicates the overlap between machine translation and ground truth reference translation) of 0.94~\cite{Papineni2002}. To the best of our knowledge, ours is the first work demonstrating the feasibility of a neural instruction combiner built into a full-fledged compiler. We also outline the open challenges that still need to be addressed later in the paper. 

Similar in spirit and complementary to our work is the work on building super optimizers from program binaries ~\cite{Bansal2006}. They work by harvesting instruction sequences from binaries, enumerating their equivalent efficient target sequences by exhaustive search techniques, and creating an offline database of optimized instruction sequences. This work is limited to X86 instruction set. We point out that  NIC can in fact leverage the binary instruction sequences harvested through super optimizers as training data. This can be done by lifting up the binary instructions to LLVM IR using existing de-compilation tools~\cite{fcd}. We plan to explore this in future work.  In our case, we automatically learn a neural model for instruction combiner, modelling it as monolingual machine translation from an un-optimized instruction sequence to an optimized instruction sequence at the compiler IR level. Since our neural model operates on a distilled IR representation (which we describe in section~\ref{section:encoding}), it is possible to port our NIC to any compiler if we can provide the NIC Inputter/Outputter modules which can convert from a compiler’s IR to NIC’s encoded/distilled representation and vice versa. Unlike super optimizers which work on specific binary instruction sets, this allows wider portability across platforms. 

\section{Background}
The heart of any compiler is the optimizer which works on the compiler Intermediate Representation (IR) of the input program. A function in compiler’s IR is split up into basic blocks (BB). IC attempts to replace a source sequence of one or more instructions in a BB by an equivalent and optimized sequence of instructions (IC can also replace single instructions with equivalent but optimized instructions). IC transformations typically include algebraic simplifications, instruction canonicalization, local constant propagation, constant folding etc. 

\begin{table*}[htb]
  \caption{Example IC transformation}
  \centering
  \begin{tabular}{|L |L|}
    \toprule
    Source Sequence & Target Sequence  \\
    \midrule
    I1:  \%sub103 = sub\: i32\: \%arg1,\: \%load1\: & I1:  \%sub103\: =\: sub\: i32\: \%arg1,\: \%load1\: \\
    I2: \%div104 = udiv\: i32\: \%sub103,\: 2\:  & I2: \%div104 = ashr\: i32\: \%sub103,\: 1\:  \\
    I3:  \%sub105 = sub\: i32\: \%gvar1,\: \%div104\: & I3:  \%sub105 = sub\: i32\: \%gvar1,\: \%div104\: \\
     \bottomrule
  \end{tabular}
  \label{tab:example}
  \end{table*}

In the IC pass the instruction sequence is scanned against multiple pattern-matching rules and once a sequence which matches the pattern is found, an equivalent and efficient transformed sequence of the identified pattern is applied to generate the new instruction sequence that replaces the original sequence.  Table~\ref{tab:example} shows an example instruction source instruction sequence wherein IC replaces a divide (UDIV) by power of 2 with shift right (ASHR) in the target instruction sequence since shifts are inexpensive than divide instruction. 

Maintenance efforts in the IC pass typically occur in pattern matching part of the pass~\cite{Zhou2021}. Once a source pattern itself has been matched correctly to determine the corresponding equivalent target sequence, generating the new sequence by up-dating the source instruction sequence is a straightforward task. Huge software code costs associated with IC and the fact that it is ubiquitous in all compilers opens the possibility of replacing the hand-coded rule driven pattern matching IC pass with a machine learnable IC pass. We attempt to do this by building a Neural Instruction Combiner (NIC) pass built on the LLVM compiler framework. We model this as a neural machine translation task from un-optimized IR sequence to an optimized IR sequence for each BB using Seq2Seq models~\cite{Bahdanau2014}. We provide a brief overview of Seq2Seq models in section~\ref{section:seq2seq}.

\label{section:encoding}
\begin{table*}[hbt]
  \caption{Example Encoding}
  \centering
  \begin{tabular}{|L |L|}
    \toprule
    Instruction Sequence & Encoding \\
    \midrule
    I1:  \%sub103 = sub\: i32\: \%arg1,\: \%load1\: & I1:\: SUB\: i32\: \%ARG1\: LOAD \\
    I2: \%div104 = udiv\: i32\: \%sub103,\: 2 & I2:\: UDIV\: i32\: SUB\: (I1)\: CONST\: 2 \\
    I3:  \%sub105 = sub\: i32\: \%gvar1,\: \%div104\: & I3: SUB\: i32\: \%GVAR1\: UDIV\: (I2) \\
    \bottomrule
  \end{tabular}
  \label{tab:encoding}
\end{table*}
In our case the input and the output sentences both belong to the same language of LLVM compiler IR and we model the problem as a monolingual machine translation task. This brings up the following open questions:
\begin{itemize}
    \item 	What should be the input sentence representation and output sentence representation for the machine learnt model? 
    \item How can we find/build a dataset for this task? 
    \item How do we integrate a machine learnt IC module into the overall optimizer pipeline? 
    \item Should the neural model directly generate the final LLVM IR sequence or have the machine learnt model output be validated and then applied on the source sequence by the optimizer?
    \item How do we validate that the IR generated from NIC is correct for downstream optimization passes?
\end{itemize}
We attempt to answer these in the next section.

\section{Description}
NIC consists of three major components 
\begin{itemize}
\item \textbf{NIC Inputter} - This is a compiler module (not an ML model) which creates a distilled representation of the IR instruction sequence corresponding to each BB. 
\item \textbf {NIC Converter} - This is a machine learnt model which takes as input, an encoded representation of IR instruction sequence corresponding to each BB of a function and converts it to an equivalent optimized sequence. This model is trained offline and employed in inference mode in LLVM optimizer.
\item \textbf{NIC Outputter} - This is a compiler module which takes as input the optimized sequence generated by NIC converter module and the original list of instruction corresponding to that BB. It recreates the standard LLVM IR optimized instruction sequence for that BB. The output of NIC Outputter is then passed to the other downstream optimization passes.
\end {itemize}
\subsection{NIC Inputter}
The input to the NIC Inputter is a regular full-fledged LLVM IR instruction stream corresponding to each BB of a function. NIC Inputter then creates a compressed encoded representation of the full LLVM IR instruction stream for that BB.  An LLVM IR instruction contains information such as debug information, named variables, initializers, instruction level completers and metadata, which is not needed by IC itself. Selection of equivalent optimized instruction sequence by IC pass is dependent on the specific operation (opcode) carried out by the instruction along with the operands of the instruction. 

IC typically does not require the other information contained in a standard LLVM IR instruction. This led us to create an encoded distilled representation of the full-fledged LLVM IR instruction sequence as input to the NIC converter module. Using the distilled representation of the LLVM IR enables our NIC Converter to deal with a smaller vocabulary and makes it more efficient. Table~\ref{tab:encoding} shows an example output from the NIC Inputter wherein the column ‘Encoding’ contains the encoded output from NIC Inputter for the LLVM IR sequence given in the column 'Instructon Sequence'. 

As seen in Table~\ref{tab:encoding}, for each LLVM IR instruction, the distilled representation contains the target opcode, its type, each of the source operands and their types. In case a source operand is being produced by an instruction, the source operand is represented by its opcode. If it is a constant, we represent it by ‘CONST’ followed by the constant value if known or by UNKVAL if unknown. In case the source operand is an incoming argument to the function or a global variable, we represent it by ARG or GVAR, respectively. We concatenate the distilled encoded instructions for the BB and this becomes the input to NIC Converter.

\subsection{NIC Converter}
\begin{figure*}[htb]
  \centering
   \includegraphics[width=0.9\textwidth]{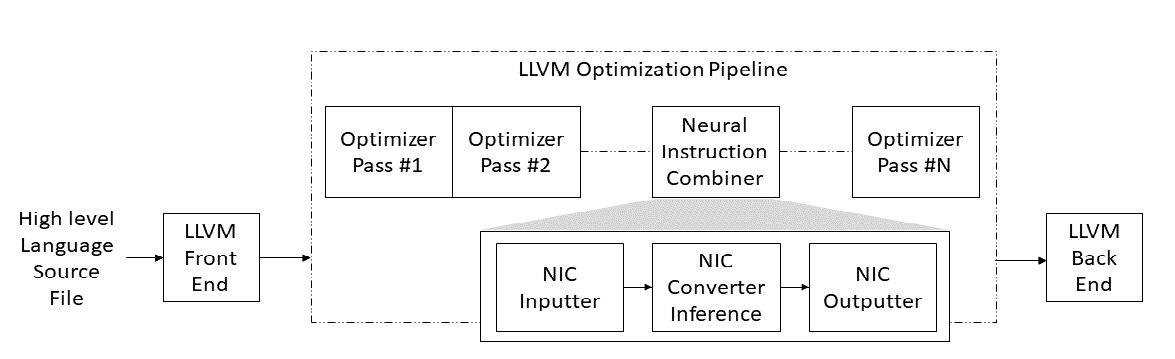}
  \caption{NIC Converter Inference Path}
  \label{fig:inference}
\end{figure*}

NIC Converter is a Sequence-to-Sequence (Seq2Seq) model which is modelled like a monolingual neural machine translator. The model is trained offline and is invoked in inference mode, during the LLVM optimizer pipeline. Figure~\ref{fig:inference} shows the components of NIC that are invoked during the inference path. We describe the training of NIC Converter later in section~\ref{section:training}.

NIC converter takes as test input, the encoded BB instruction sequence from NIC Inputter, and predicts an optimized encoded instruction sequence corresponding to it. NIC Converter consists of a Seq2Seq model internally. The predicted encoded instruction sequence is then fed to the NIC Outputter. The source sentence is the encoded instruction sequence corresponding to a BB and the target sentence is the optimized sequence for that BB. We next briefly provide a brief description of Seq2Seq models used in our NIC Converter. 
\subsubsection{Sequence to Sequence Models}
\label{section:seq2seq}
\begin{figure*}[htb]
  \centering
   \includegraphics[width=0.8\textwidth]{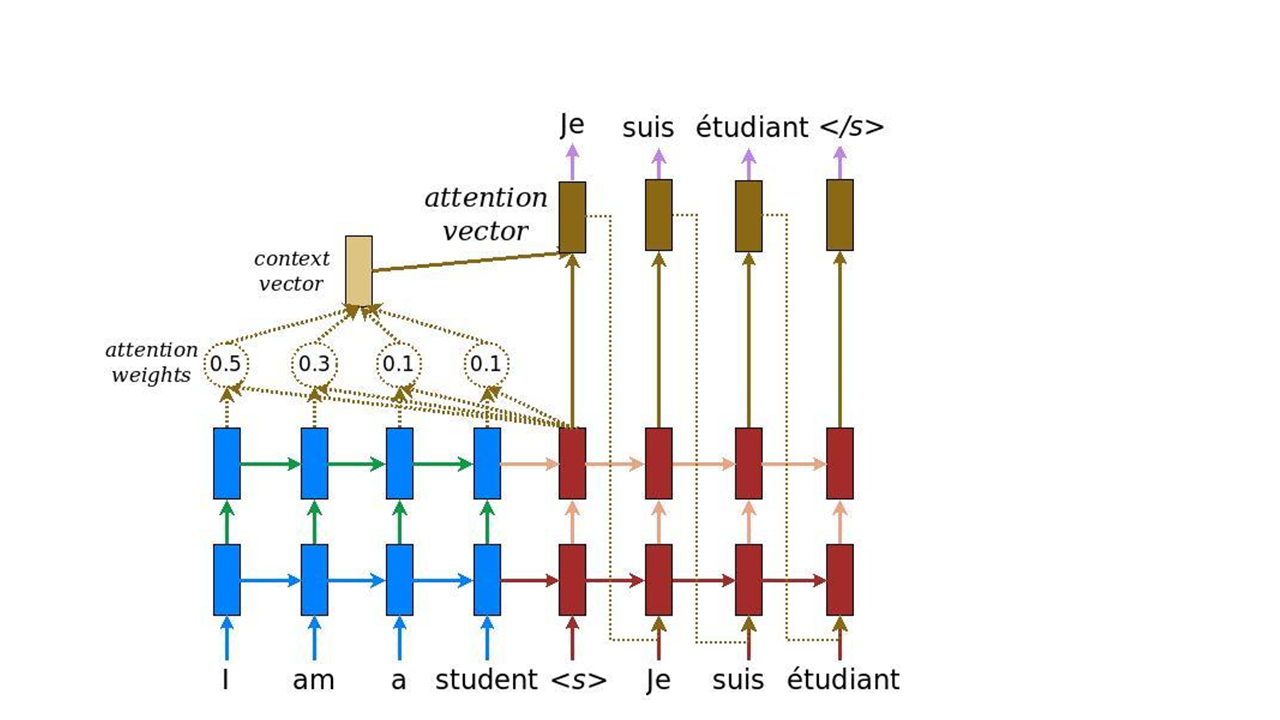}
  \caption{Seq2Seq Model deployed for Neural Machine Translation Task}
  \label{fig:seq2seq}
\end{figure*}

Figure~\ref{fig:seq2seq} shows an example Seq2Seq model for neural machine translation. The Seq2Seq model consists of two subnetworks, the encoder and the decoder. The encoder and decoder are typically either Recurrent Neural Networks (RNN) of the form Long Short Term Memory (LSTM) networks~\cite{lstm} or they can be transformer blocks~\cite{Vaswani2017}. The input sentence in source language is represented through token embeddings of the constituent tokens/words.  The encoder receives the input token embedding sequence  and produces a compact/encoded representation of the input sequence, trying to summarize or condense all of its information. In the basic Seq2Seq model, the encoder’s last cell output is fed to the decoder as initial state. At each time step, the decoder generates an output token based on the previous output token and its current state, as well as updating its own state for the next time step. 

In case of RNN based Seq2Seq models, the encoder and decoder typically consists of fixed length sequence of LSTM cells (the number of cells in encoder/decoder can be different).  Encoder and decoder can also consist of stacked layers of LSTM cells instead of a single layer of LSTM cells. The encoder layers can be either unidirectional or bidirectional. Decoder layers are unidirectional. In the case of neural machine translation, Seq2Seq model is used to translate an input sentence in one language to a target sentence in another language. As shown in the example in Figure~\ref{fig:seq2seq}, given an input sentence in English, the words in the sentence are represented using their word embeddings, and fed to the encoder. The encoder output is passed to the decoder which generates each word of the target French sentence.  

In the case of NIC Converter, the source sentence is the distilled representation of the unoptimized IR instruction sequence at the input of the Instruction Combiner (which is the output from NIC Inputter). This input sentence is represented as a sequence of token embeddings (the embeddings are task specific and trained as part of the Seq2Seq model training). It is then passed through the encoder-decoder network of the Seq2Seq model. The output sequence from the decoder is the predicted optimized instruction sequence corresponding to the input sequence. This becomes the output of the NIC Converter.
\begin{figure*}[htb]
  \centering
   \includegraphics[width=0.8\textwidth]{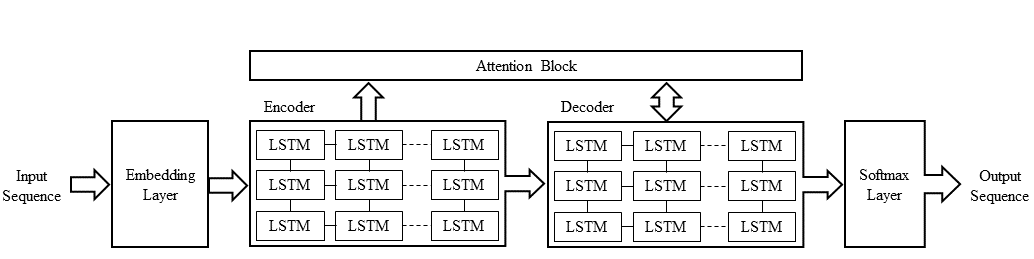}
  \caption{RNN based NIC Converter Seq2Seq model}
  \label{fig:lstm}
\end{figure*}
Since the encoder compresses the information contained in the entire input sequence into a single output vector from its last stage output (typically known as context vector), the basic seq2seq models have difficulty in retaining all the information from long input sequences. Attention mechanism was proposed to address this issue~\cite{Bahdanau2014}. Instead of having the context vector taken from the last cell of the encoder which can result in lossy compression of the information contained in the input sequence, attention mechanism is used to create a context vector which is a weighted combination of the outputs from all the cells of the encoder at each step of the decoding process. 

At every decoding step, the decoder will be informed how much “attention” needs to be paid to each input word using a set of attention weights. These attention weights provide contextual information to the decoder for generation of each output token. All hidden states of the encoder are used to generate the context vector, weighing each state by its corresponding attention weight. A detailed mathematical representation of attention mechanism can be found in ~\cite{Bahdanau2014}.     

As is typical practice, our Seq2Seq models are based on the standard encoder-decoder framework with attention ~\cite{Sutskever2014}. We consider two design choices for the encoder-decoder network, one based on RNN with a single head attention ~\cite{Bahdanau2014}, and another based on standard transformer model with multi-head attention ~\cite{Vaswani2017}.  The input sentence is converted in-to fixed length representation using the encoder from which the decoder emits the target sentence, one token at a time. Attention mechanism is employed to improve the ability of the Seq2Seq model to attend to the most relevant encoder outputs, when decoding each respective token. 

A high level block diagram of our RNN based Seq2Seq model is shown in Figure~\ref{fig:lstm}. Our RNN based Seq2Seq model deploys bidirectional stacked LSTM layers for the encoder and unidirectional stacked LSTM layers for the decoder. We use the additive encoder-decoder cross attention mechanism proposed in ~\cite{Bahdanau2014} for our RNN based Seq2Seq model, wherein the context vector is a weighted composition of encoder outputs by the attention weights. Attention weights are learnt as part of the task training. Attention block is intended to capture the soft alignment between the tokens in the source sentence to the tokens in the target sentence and the learnt alignment scores are used to compute the context vector for the decoder. The attention score between the 'ith' token in target sentence to the 'jth' token in the source sentence is computed as a function of decoder's hidden state for the previously emitted token and the encoder's output for the current input token. In lay terms, higher the influence of an input token 'j' on the output token 'i', greater would be its corresponding attention score. Typically the attention block consists of a feed forward layer followed by a softmax layer~\cite{Bahdanau2014}. 

\begin{figure}[htb]
  \centering
   \includegraphics[width=0.5\textwidth]{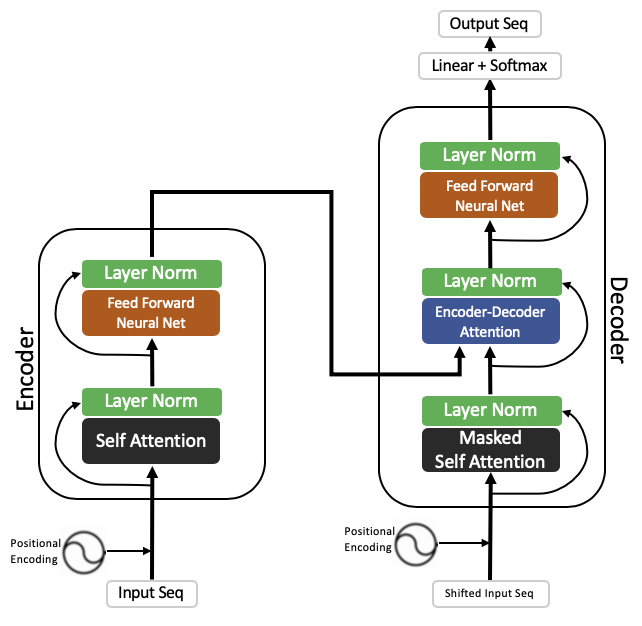}
  \caption{Transformer based NIC Converter Seq2Seq model}
  \label{fig:transformer}
\end{figure}
A high level block diagram of our transformer based Seq2Seq model is shown in Figure~\ref{fig:transformer}. Instead of LSTM cells, Transformer encoder consists of layers of transformers. Each transformer encoder layer consists of two sub-layers namely (a) multi-head self attention layer and (b) point-wise feed forward layer, with each followed by a layer normalization sub-layer(LayerNorm) as shown in Figure~\ref{fig:transformer}~\cite{layernorm}. While RNN based Seq2Seq model deploy cross-attention (also known as inter-attention) only, transformer based Seq2Seq models deploy both self attention (known as intra attention) and cross-attention~\cite{Vaswani2017}. Self-attention is an attention mechanism relating different positions of a single sequence in order to compute a representation of the sequence. Basically Self attention is applied on the tokens of the same sequence and hence it is known as intra-attention. Transformer decoder layers are similar to transformer encoder layers except that they also have in addition,  a cross-attention decoder layer as seen in Figure~\ref{fig:transformer}. As part of our evaluation, We experiment with varying number of transformer layers, number of attention heads and feed forward layer width for our NIC Converter and report the results in Section~\ref{section:exp_results}. 
\subsubsection{Training of NIC Converter}
\label{section:training}
\begin{figure*}[htb]
  \centering
   \includegraphics[width=0.8\textwidth]{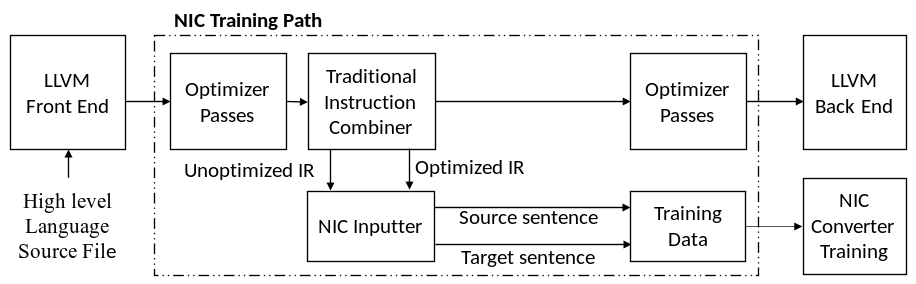}
  \caption{NIC Converter Training path}
  \label{fig:training}
\end{figure*}

For the current work, we train NIC Converter in a supervised manner using data obtained from traditional IC similar to behavior cloning~\cite{behavior}. As shown in Figure~\ref{fig:training}, training data is generated in an offline phase by the compiler using the NIC Inputter module. Given a source file in high level language supported by the compiler, the compiler takes the input one function at a time and generates the BB level IR instruction sequences corresponding to each BB in the function. The compiler invokes the NIC Inputter on these unoptimized instruction sequences to obtain the encoded source sequences and then passes the original (unencoded) instruction sequences through the traditional (non-neural) IC phase. The optimized instruction sequences at the output of the traditional IC phase are then passed through NIC Inputter and encoded target sequences are obtained. 
The compiler maintains the BB level mapping between the unoptimized and optimized encoded sentence pairs and creates the list of sentence pair $<$unoptimized encoded instruction sequence, IC optimized encoded instruction sequence$>$ corresponding to each BB. This is done at each function level.  Given a set of source files in a high-level language, this process is repeated for all source files, and a list of sentence pairs are generated as training data for the NIC Converter by the compiler. We then train the Seq2Seq model  to create the machine learnt NIC convert-er model, using the standard cross-entropy loss objective at each token level. We design and implement a variant of standard Seq2Seq model wherein we use compiler knowledge to guide the attention process, which we describe next.
\subsubsection{Compiler Guided Attention Training}
Attention enables the decoder to selectively consider relevant words of the source sentence when emitting each to-ken in the target sentence.  Standard attention is learnt with the indirect objective of improving the translation quality and is not learnt in a supervised manner with respect to word/phrase alignment between source/target sentence since direct word alignment information is not typically available for the training data in general. Hence it may not always correlate well with the alignment between source and target sentences.  However, in case of NIC Converter, since the compiler has exact knowledge of which source instructions are responsible for generating the corresponding target instructions at each constituent token level, we leverage the compiler knowledge in improving the soft alignments learnt by the attention network.

During training data generation, a compiler guided attention matrix CA is created by the compiler for each BB. CA matrix terms are fixed attention scores provided by the compiler and are not learnt during training. For each source sentence, compiler has information of which source instruction tokens map to corresponding optimized target instruction tokens and uses this to set each element of the CA matrix. Each element CA[i, j] corresponds to the probability of whether the ih token in target sentence (of length T) is mapped to jth token in the source sentence and the total probability of one is distributed among the relevant mapped tokens while the non-relevant mapped tokens are set to zero. This is like hard attention with CA[i, j] being non-zero if target token ‘i’ is mapped to source instruction token ‘j’ by the compiler and else zero. 

We smooth this hard attention matrix with a small correction term delta, adding it to all zero terms and adjusting non-zero terms accordingly to maintain the row sum as 1. CA matrix has similar semantics and same dimension as the learnt attention matrix ‘A’ with standard attention. In case of single head attention, A is the single head cross attention weights, and in MHA, we use the decoder cross-attention weights of last layer’s head 0 as the learnt attention matrix ‘A’.

We use CA matrix to force the learnt attention weights ‘A’ to be closer to it during the training process by adding an additional loss term to the training objective. This is compiler attention mismatch loss term which is the divergence of the learnt attention weights in each training step from the compiler guided attention matrix CA. We model the Compiler Attention Mismatch (CAM) loss between CA and A matrices using the standard cross entropy loss function as below:

\begin{align}
CAM\; Loss (CA,A) = -(1/T)\sum_{i}\sum_{j} CA[i][j] * log(A[i][j]) \\
Total\; Loss = Decoder\; Cross\; Entropy\; Loss \; + \; CAM \;Loss 
\end{align}

Compiler guided attention is not used during inference, as we hypothesize that compiler guided attention will enable the model to learn the appropriate attention weights during training itself. 

\subsection{NIC Outputter}
The NIC Outputter takes two inputs:
\begin{itemize}
    \item the predicted instruction sequence from the NIC Converter 
    \item the original unmodified full-fledged LLVM IR instruction stream corresponding to that BB.
\end{itemize}

Given that NIC Converter predicts the most probable target instruction sequence, we also enforce specific checks in the NIC Outputter to ensure that the generated target sequence does not violate the compiler integrity checks. NIC Outputter performs a verification check first on the generated target sequence. It checks that for each instruction in the generated sequence, the number of operands corresponding to that opcode are correct, and that each operand has previously been defined in the generated instruction sequence, and that the last instruction in the sequence is a branch/return instruction. In case any of these conditions are violated, it discards the generated target instruction sequence and outputs the source instruction stream corresponding to that BB as is.

If the verification checks on the generated sequence are satisfied, then NIC Outputter takes the generated instruction sequence and applies it on the source instruction stream corresponding to that BB to produce the transformed full LLVM IR instruction stream. If NIC Converter predicted output is the same as its input, (the NIC Converter has not found any suitable transformation for the given input sequence) NIC Outputter just reproduces the unmodified source IR sequence as is without any checks. NIC Outputter also invokes the LLVM function level verification. This checks that the CFG (Control Flow Graph) is valid, all instructions are associated with a BB and specific instruction level checking based on the instruction type (the types of operands of binary operator are of the same type, the Static Single Assignment ~\cite{Cytron1989} form is valid, shifts and logicals happen only on integral types etc). These checks serve to ensure that the generated IR is valid. 

We then check for translation validity of the generated sequence by passing it through a well-known LLVM IR translation validity checker ALIVE2~\cite{alive22021}. ALIVE2 is a fully automatic bounded translation validation tool for LLVM that supports all of its forms of undefined behavior. ALIVE2 checks pairs of instruction sequences in LLVM IR for refinement using an SMT solver. A refinement relation is satisfied when, for every possible input state, a target sequence displays a subset of the behavior of the source sequence. In the absence of undefined behaviors, refinement degenerates to simple equivalence. Checking for translational validity between source and target sequences ensures semantic equivalence between source and target instruction sequences~\cite{alive22021}. ALIVE2 does not require any changes in LLVM Compiler. While ALIVE2 can check for any intra-procedural optimization validity in LLVM, in our case, we use it to verify the translation validity of the NIC generated seqquences which are limited to basic block level local code rewriting transformations. Instruction sequences rejected by ALIVE2 as non-valid are rejected at the NIC Outputter. 

\section{Experimental Evaluation}
\begin{table*}[!htb]
  \caption{NIC Experimental Results}
  \centering
  \begin{tabular}{|T|c|c|c|c|c|c|c|c|c|}
  \hline
  Model & Bleu & Rouge-1 r & Rouge-1 p & Rouge-2 r & Rouge-2 p& Rouge-l r & Rouge-l p &  EM(unopt) & EM(opt) \\
  \hline
  Bidirectional LSTM: $3$ layer encoder, unidirectional greedy decoder & 0.93 & 0.98& 0.90& 0.96& 0.91& 0.97 & 0.93 & 0.93& 0.68 \\
  \hline
  Transformer with layers=$4$, dmodel=$128$,  dff=$512$, heads=$8$ & 0.94& 0.98& 0.90& 0.96 & 0.91 & 0.97 & 0.94 & \textbf{0.94} & \textbf{0.72} \\
  \hline
  Transformer with layers=$6$, dmodel=$512$,   dff=$2048$, heads=$8$ &
  0.91& 0.98& 0.90& 0.96 & 0.91 & 0.96& 0.93 & 0.93 & 0.71 \\
  \hline
  Transformer with layers=$2$, dmodel=$512$,   dff=$2048$, heads=$8$ &
  0.93 & 0.98& 0.90& 0.96& 0.91 & 0.97& 0.94 & 0.93 & 0.70 \\
  \hline
  Transformer layers=$4$, dmodel=$128$,  dff=$512$, heads=$8$ and no POS Emb & 0.94 & 0.98& 0.90& 0.96& 0.92 & 0.97& 0.94 & 0.94 & 0.70 \\
  \hline
  Transformer with layers=$4$, dmodel=$512$,   dff=$2048$, heads=$16$ &
  0.93 & 0.98& 0.90& 0.96& 0.91 & 0.97& 0.93 & 0.94 & 0.71 \\
  \hline
  Bi-directional LSTM (3 layer encoder, 1 layer decoder) with guided compiler attention & 0.93 & 0.98& 0.89& 0.96& 0.91 & 0.96& 0.94 & 0.93 & \textbf{0.70} \\
  \hline
  Transformer with layers=$4$, dmodel=$128$,  dff=$512$, heads=$8$ with compiler guided attention & 0.94 & 0.98& 0.90& 0.96& 0.92 & 0.97& 0.93 & 0.94 & 0.72 \\
  \hline
  \end{tabular}
  \label{tab:results}
\end{table*}

\subsection{Dataset Description}
Given the novelty of our task, there are no readily available datasets which can be used to train the NIC Converter model. Hence, we build a new dataset for this task and plan to make it available publicly. The dataset consists of sentence pairs both from the same language of LLVM IR. The source sentence is the encoded distilled sequence of the LLVM IR instruction sequence of a BB at the input of traditional IC Pass. The target sentence is the encoded distilled sequence of that instruction sequence at the output of the traditional IC Pass.  

As part of the dataset, we provide along with these sentence pairs of encoded basic block IR sequences, the full-fledged LLVM instruction sequence at the input and output of traditional IC Pass. The full-fledged IR sequences are just for informational purpose and are not used during training/inference of the NIC converter model.  The training data is generated by invoking our modified version of LLVM compiler pipeline on the C/C++ application source files. NIC inputter is the only component invoked during training data generation.  The full-fledged instruction sequence corresponding to each BB at input and output of the traditional IC Pass are passed through the NIC inputter to generate the training data. Our NIC Converter Seq2Seq model is trained in a supervised manner using this dataset.

Training data is generated by invoking our modified version of LLVM compiler pipeline on the C/C++ application source files.  For generating the training data, we used a collection of C/C++ source files from LLVM Application Test Suite and from Angha Bench Test Suite ~\cite{daSilva2021}. LLVM application test suite consists of code fragments as well as large complete programs. Angha Bench consists of more than a million of C code snippets extracted from various open source repositories and made compilable by automatically adding the requisite type declarations. This results in 759K sentence pairs. 

We note that two different basic blocks instruction sequences when converted to the distilled sequence, can end up with the same source sentence due to the compressed representation. Hence we dedup our training data generated so that each source sentence is unique from all others in the dataset.  The final deduped dataset size is $367K$ and contains 66\% of unoptimized sequences (where source and target sentences are identical) and 34\% optimized sequences (where source and target sentences are different). Given the imbalanced nature of the dataset (with 34\% of sentences are optimized), we use the Exact Match percentage on optimized sequences as our primary task evaluation metric. We make our code and dataset available for review as part of the artefact submission. 

\subsection{Experimental Results}
\label{section:exp_results}
We evaluate several Seq2Seq models for NIC Converter as shown in Column $1$ of Table ~\ref{tab:results}. The two main network choices are RNN and the standard transformer models with greedy decoder (we did not see any significant change with beam decoder). 
For transformer models, we experiment with different settings for the number of layers, attention heads, embedding size dimension (denoted as dmodel) and feed forward layer width (denoted as dff) in Table~\ref{tab:results}.  We implemented various models in TensorFlow Python framework for evaluation.  All parameters are initialized by uniform distribution over [-0:1; 0:1]. The mini-batch stochastic gradient descent algorithm is employed to train the model with batch size of 64 and number of training epochs being 10. Hyperparameters were chosen based on experimentation with validation set. In addition, we use Adam optimizer with custom rate scheduler ~\cite{Vaswani2017}. We do a train/validation/test data split of 90\%, 5\% and 5\% respectively. We used the LLVM Compiler release 9 version.

\begin{figure}
  \centering
   \includegraphics[width=0.5\textwidth]{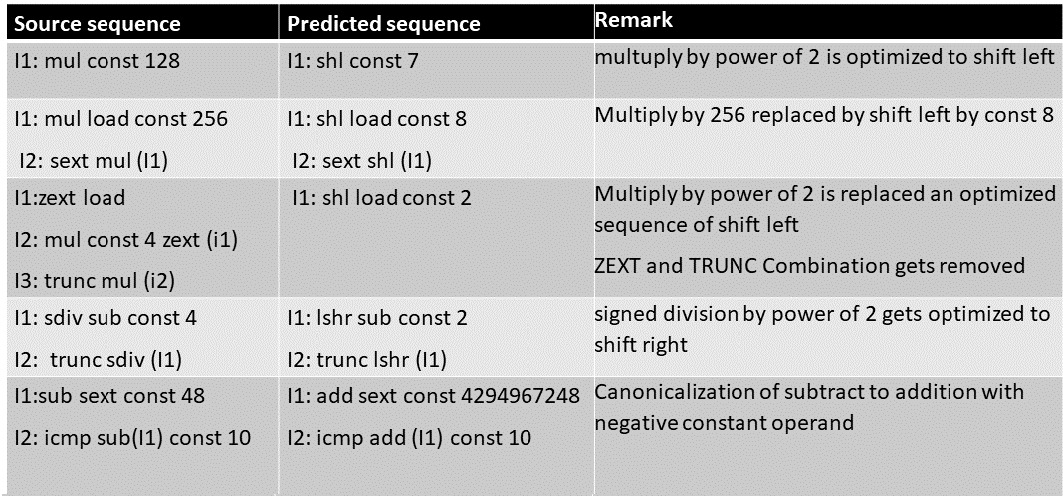}
  \caption{Example Outputs from NIC Converter}
  \label{fig:example_outputs}
\end{figure}
For NIC Converter inference performance, we report the standard NMT evaluation metrics of Bleu precision ~\cite{Papineni2002} and Rouge scores ~\cite{Lin2004} in Table~\ref{tab:results}. However, our task being code optimization, requires generation of exact encoded representation (even a single wrong token  will lead to incorrect code). \textit{ our task specific metric is comparison of Exact Match (EM) results for the entire instruction sequence for each BB between the predicted sequence and the ground truth.} We show the EM results separately in last $2$ columns of Table~\ref{tab:results} respectively for (a) EM(opt) where the ground truth is an optimized translation of the input sequence and (b) EM(un-opt) where the ground truth is same as the input sequence.

BLEU (bilingual evaluation understudy) is an algorithm for evaluating the quality of text which has been machine-translated from one language to another. Quality of translation is a mesaure of how much the machine translation is similar to a human reference translation. BLEU's output is always a number between 0 and 1 with a value equal to 1 indicating that machine translation and human translation were identical. Bleu precision is typically evaluated at multiple n-gram level with average across all n-gram levels being reported as a single final score. A detailed definition of Bleu precision can be found in ~\cite{Papineni2002}. Rouge score is another well known set of metrics for evaluating the quality of machine translation. Rouge stands for Recall-Oriented Understudy for Gisting Evaluation~\cite{Lin2004}. Rouge-n score represents the n-gram overlap between the machine generated and ground truth reference translations. While originally Rouge was intended as a recall measure, it has been augmented with precision scores as well~\cite{Lin2004}.We report both Rouge precision (denoted as Rouge-* p) and recall (Rouge-* r) scores for Rouge-1, Rouge-2, Rouge-l n-gram overlaps in Table~~\ref{tab:results}. 

Across all models, Bleu and Rouge scores are not significantly different, indicating their general translation capability (there were differences only beyond last 2 digits). As is expected, the EM percentage is much higher for the unoptimized sequences and is an indicator of model ability to reproduce the exact input sequence correctly. In case of optimized sequences, model’s Exact Match is around 69\%-73\% indicating considerable room for further improvements. We find that transformer model with 8 attention heads, 4 layers, and embedding dimension 128 has the best Exact Match results in our experiments.

We find in general that transformer models are better than RNN models by 2-3\% of EM percentage. While compiler guided attention improved EM percentage for RNN model by 2\%, it is still lower than that of MHA transformer models. Compiler guided attention had negligible impact on transformer models. We hypothesize that this may be due to MHA’s ability to better capture word alignment information in multiple head attention weights compared to single head attention. We find that removing transformer positional encodings did not have any impact on model performance, similar to earlier works ~\cite{Clouatre2021}. We did not see any improvement in performance by increasing the number of layers and feed forward layer dimensions in our transformer models. The compile time impact was negligible due to the deployment of NIC Converter in inference mode and the verification checks in LLVM Compiler pipeline for our test dataset evaluation. 

\subsection{Exact Match Results Analysis}
 \begin{table}[htb]
  \caption{Exact Match Error Analysis}
  \centering
  \begin{tabular}{|c|c|}
    \hline
    Type of Error    & Occurrence    \\
    \hline
    Incorrect Constant & 42.3\%       \\
    \hline
    Opcode Mismatch     &  34.9\%       \\
    \hline
    Type Issue (Sext/Zext)     &    6.7\%      \\
    \hline
    Operand Mismatch           &    1.4\%      \\
    \hline
    Others                     &    14.7\%     \\
    \hline
  \end{tabular}
  \label{tab:subtab2}
\end{table}

We analyzed NIC Converter outputs to understand the model's strengths and shortcomings. Figure~\ref{fig:example_outputs} shows some samples of the NIC Converter generated sequences. In case of optimized sequences, we find that NIC not only generates the optimized opcode, but also correctly fixes up the uses of the replaced opcode with the newly generated opcode, allowing us to hypothesize that the model has learnt the implicit use-def chains~\cite{Cytron1989}in the encoded representation. 

Table~\ref{tab:subtab2} shows the major reasons for exact match errors by NIC. One of the common mistakes the model exhibited was in generating correct values for synthesized constants for certain operations. A frequent LLVM IR instruction is the GetElementPtr (GEP) instruction used to get the ad-dress of a sub element of an aggregate data structure such as a ‘struct’ or ‘array’. For nested aggregate data structures, a sequence of GEP operations is emitted with appropriate constants from base address. One of the IC optimization sequences is coalescing multiple GEP operations into a single GEP with modified constant indices as operands. We found that NIC had issues in synthesizing the GEP indices correctly. We reason that NIC is not able to learn the rules to compute the GEP offsets correctly based on individual GEP operations. It ends up reproducing the memorized frequent constant values as GEP indices generating erroneous sequence.  

A similar problem was noticed in the case of ‘Alloca’ instruction sequence optimization where stack offsets constant values were generated incorrectly by NIC. These errors are caught during the verification of the generated code sequence in the NIC Outputter, dropping the suggested replacement from NIC from being applied. For frequently occurring/unique constants such as powers of two occurring in Shift instructions, the model outputs the correct constants both in optimized and unoptimized sequences. However, for arbitrary constants such as those occurring in GEP/Alloca operands, model ends up making mistakes. The error analysis indicates that our current model does not handle synthesized constants in the instruction sequence well. This needs to be addressed in future work. Other errors include wrong OPCODE generation in predicted sequence, unnecessary operand swap canocalization and incorrect type extension elimination.  These errors can be addressed by expanding the training dataset used to train the NIC converter and by deploying contrastive learning during NIC converter training~\cite{contrastive}.  We plan to explore this as part of future work.
\section{Related Work}
Of late, there has been considerable interest in applying deep learning techniques to compilers in the areas of phase ordering ~\cite{Huang2019}, selection of optimization heuristics  ~\cite{Cummins2017} and as part of optimization itself in register allocation ~\cite{Das2020} and inlining ~\cite{Troffin2021}. Machine learnt models have been used in optimization heuristics selection such as prediction of unroll factors ~\cite{Stephenson2005}, inlining decisions ~\cite{Simon2013}, vectorization~\cite{HajAli2020}, ~\cite{Mendis2019} etc.  

Our work falls under the category of deploying ML models directly in compiler optimizations. Similar in spirit and complementary to our work, there has been work on building super optimizers by creating a database of possible optimized sequences from the binaries ~\cite{Bansal2006}. These techniques work by harvesting instruction sequences from binaries, enumerating their equivalent efficient target sequences by exhaustive search techniques, and creating an offline database of optimized instruction sequences which can then be looked up for a given sequence of instructions. These methods incur high overheads due to huge candidate search space, and equivalence checking via approximate testing of selected sequences, making them difficult to deploy. This line of work is limited to X86 instructions.  

There has been work on improving the brute force search for optimization sequences ~\cite{Schkufza2012} using random search and RL methods ~\cite{Bunel2017}. NIC can leverage the optimized instruction sequences generated by super-optimizers by lifting them to LLVM IR using decompilation techniques~\cite{fcd} (We plan to explore this in future work) and then using them for training the NIC Converter making them complementary to our work. This will enable expanding the NIC training dataset considerably. Since our neural model operates on an encoded condensed IR representation, it is possible to port our NIC to any compiler if we can provide the NIC Inputter/Outputter modules which can convert from compiler’s IR to NIC’s encoded representation and vice versa. Unlike super optimizers which work on specific binary instruction sets, this allows wider portability. 

\section {Open Issues and Conclusion}
In this paper, we explored the feasibility of replacing the traditional Instruction Combiner with a neural instruction combiner in a widely used production level compiler. We find that we were able to train a neural instruction combiner module and integrate it with LLVM compiler’s optimizer pipeline. However there are still open issues. Our current work leaves considerable optimization potential on the table (only $72$\% of optimization opportunities are realized by NIC). Validity checks of generated sequence currently include compiler driven IR Validity checks and translation validity checking using ALIVE2 external tool~\cite{alive22021}. Leveraging scalable automatic post-editing~\cite{ape2020} and program repair techniques~\cite{sed2020} can help improve validity checks in NIC. We also plan to expand NIC training data with IR instruction sequences that can be leveraged from existing super-optimizer databases. These need to be addressed for a robust NIC deployment in production compilers. 
\bibliography{sample2}
\end{document}